\documentclass[conference]{IEEEtran}
\usepackage{cite}
\usepackage{amsmath,amssymb,amsfonts}
\usepackage{algorithmic}
\usepackage{graphicx}
\usepackage{textcomp}
\usepackage{xcolor}
\usepackage{subcaption}
\usepackage{smile}
\usepackage{bbding}
\usepackage{hyperref}
\def\BibTeX{{\rm B\kern-.05em{\sc i\kern-.025em b}\kern-.08em
    T\kern-.1667em\lower.7ex\hbox{E}\kern-.125emX}}
\begin{document}

\title{Making Reconstruction-based Method Great Again for Video Anomaly Detection
}

\author{\IEEEauthorblockN{Yizhou Wang,
Can Qin, Yue Bai, Yi Xu, Xu Ma and Yun Fu}
\IEEEauthorblockA{
Northeastern University, Boston, USA}
wyzjack990122@gmail.com, \{qin.ca, bai.yue, xu.yi, ma.xu1\}@northeastern.edu, yunfu@ece.neu.edu}

\maketitle

\begin{abstract}
    Anomaly detection in videos is a significant yet challenging problem. Previous approaches based on deep neural networks employ either reconstruction-based or prediction-based approaches. Nevertheless, existing reconstruction-based methods {\bf 1)} rely on old-fashioned convolutional autoencoders and are poor at modeling temporal dependency; {\bf 2)} are prone to overfit the training samples, leading to indistinguishable reconstruction errors of normal and abnormal frames during the inference phase. To address such issues, firstly, we get inspiration from transformer and propose {\bf S}patio-{\bf T}emporal {\bf A}uto-{\bf T}rans-{\bf E}ncoder, dubbed as STATE, as a new autoencoder model for enhanced consecutive frame reconstruction. Our STATE is equipped with a specifically designed learnable convolutional attention module for efficient temporal learning and reasoning. Secondly, we put forward a novel reconstruction-based input perturbation technique during testing to further differentiate anomalous frames. With the same perturbation magnitude, the testing reconstruction error of the normal frames lowers more than that of the abnormal frames, which contributes to mitigating the overfitting problem of reconstruction. Owing to the high relevance of the frame abnormality and the objects in the frame, we conduct object-level reconstruction using both the raw frame and the corresponding optical flow patches. Finally, the anomaly score is designed based on the combination of the raw and motion reconstruction errors using perturbed inputs. Extensive experiments on benchmark video anomaly detection datasets demonstrate that our approach outperforms previous reconstruction-based methods by a notable margin, and achieves state-of-the-art anomaly detection performance consistently. The code is available at \url{https://github.com/wyzjack/MRMGA4VAD}.
\end{abstract}


\section{Introduction}
Anomaly detection is a significant yet challenging field in machine learning due to the difficulty of modeling unseen anomalies~\cite{wang2022self,wang2021sla}. Video anomaly detection (VAD)~\ refers to the process of identifying events that do not conform to expected behaviour~\cite{chandola2009anomaly,bai2020dual,bai2021human,Bai_Wang_Tao_Li_Fu_2021}. It has attracted significant attention from academia and industry due to its video surveillance and municipal management applications~\cite{chandola2009anomaly,ramachandra2020survey}. Meanwhile, VAD is extremely challenging.

Existing approaches fall into two main categories: traditional methods and deep-learning-based methods. Traditional methods perform classic anomaly detection techniques on top of the handcrafted features~\cite{mahadevan2010anomaly,cong2011sparse}. The inability to capture discriminative information makes these methods uncompetitive in performance, and labor-intensive feature engineering processes obstruct them from being practical and widely used. Taking advantage of deep neural networks' extraordinarily discriminative power, deep-learning-based methods have been dominating the VAD field recently. Deep-learning-based methods can further be grouped into reconstruction-based methods and prediction-based methods. Methods based on reconstruction~\cite{hasan2016learning,luo2017remembering,hochreiter1997long} extract feature representations using auto-encoders (AE) and try to reconstruct the input. It is expected that abnormal clips would have comparatively larger reconstruction errors than the normal ones. Nonetheless, existing reconstruction methods tend to concentrate on low-level pixel-wise error instead of high-level semantic features due to the poor reasoning ability along the temporal dimension and the tendency to overfit by convolutional AEs~\cite{liu2018future}. Therefore, reconstruction-based methods have recently been transcended by prediction-based methods, which predict the current frame using previous frames~\cite{zhao2017spatio,liu2018future} or adjacent frames~\cite{yu2020cloze}. As the target outputs do not exist in the inputs, prediction-based methods can better model and excavate the temporal relationship of consecutive frames. However, methods based on prediction suffer from the short length of the constructed cubes, and many of them require model ensembling to guarantee high performance~\cite{zhao2017spatio,ye2019anopcn,tang2020integrating,yu2020cloze}, making these solutions not scalable and hindering their deployment in intricate settings.

To this end, we rethink the reconstruction-based VAD methods and propose an object-level reconstruction-based framework for efficient anomaly detection in videos. Specifically, we first take advantage of a pre-trained object detection model to generate bounding boxes and extract object-centric patches. For each object-centric patch in the current frame, we scale to both the previous and the trailing frames to construct a spatio-temporal context cube, where we perform patch sequential reconstruction. We propose a novel architecture {\bf S}patio-{\bf T}emporal {\bf A}uto-{\bf T}rans-{\bf E}ncoder, which we call as \textit{STATE}, to better model the temporal relationship of the consecutive patches. we adapt the self-attention modules in the transformer into convolutional auto-encoders for both raw pixel and motion reconstruction, and we innovatively introduce a convolutional network to learn the spatio-temporal attention. The training loss is designed as the sum of the raw and motion reconstruction errors. With the spatio-temporal attention stacks, our STATE can better reconstruct during training and reason through temporal axis more efficiently during testing. Furthermore in the testing phase, we add input perturbation using the gradient of the reconstruction error w.r.t. the input frames to further reduce the testing reconstruction error. Such practice further enlarges the distribution gap of the reconstruction error of normal and abnormal testing frames. In such manner, we break through the bottleneck of the reconstruction-based VAD method and make it great again for VAD. 

\section{Related Works}
\captionsetup{width=\textwidth}
\begin{figure*}[t]
   \centering
      \includegraphics[width=\linewidth]{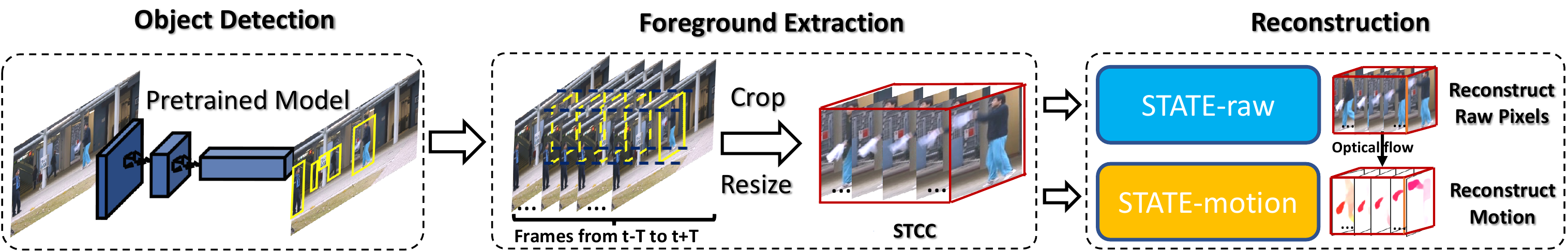}
      \caption{The pipeline of our framework: 1) Object extraction utilizing pre-trained model 2) Spatio-temporal context cube construction via extending, cropping, and resizing 3) Object-level patch sequence reconstruction with two branches.}
   \label{fig: pipeline-structure}
\end{figure*}
\subsection{Anomaly Detection in Videos}

\paragraph{Reconstruction-based Methods}
Thanks to the remarkable success of deep learning~\cite{krizhevsky2012imagenet,mittal2020hyperstar,Liu_2022_CVPR,Chen2020On,deng2022self}, reconstruction-based methods began to emerge several years ago. These approaches learn normal patterns of the training data using deep neural networks like AEs to reconstruct video frames. In the testing phase, anomalous frames are distinguished out if the reconstruction error is large. For instance,~\cite{hasan2016learning} pioneers AE-based VAD by introducing convolutional autoencoders.~\cite{luo2017remembering} plugs convolutional LSTM~\cite{hochreiter1997long} network into AE for better reconstruction of temporal video sequences.~\cite{gong2019memorizing} combines AE with memory mechanisms, and~\cite{chang2020clustering} designs k-means clustering after the encoders to ensure appearance and motion correspondence. Besides AEs, generative models such as GANs~\cite{ravanbakhsh2017abnormal} and variational autoencoders~\cite{yan2018abnormal} are also utilized for reconstruction. Even though they are contingent on the assumption that anomalous events would lead to bigger reconstruction error, this does not necessarily hold in practice due to the overfitting property of deep AEs~\cite{liu2018future}. Moreover, all the above-mentioned AE methods, even with RNN modules, are still poor in modeling temporal dependencies among video frame sequences.

\paragraph{Prediction-based Methods}
Methods based on frame prediction have been prevailing recently. They learn to generate the current frame using previous frames, and a poor prediction is treated as abnormal.~\cite{liu2018future} first proposed frame prediction method as a baseline for VAD and they choose U-Net~\cite{ronneberger2015u} as the predictor model.~\cite{lu2019future} further enhances frame prediction method performance using a convolutional variational RNN. A congenital idea is to combine and integrate prediction-based methods with reconstruction-based methods~\cite{zhao2017spatio,ye2019anopcn,tang2020integrating}, leading to the so-called hybrid methods. ~\cite{yu2020cloze} proposed to iteratively erase one frame in the temporal cube sequence and employ the rest to ``predict'' the erased one. %

\subsection{Input Perturbation in Anomaly Detection}
The notion of input perturbation in deep learning is first proposed by~\cite{goodfellow2014explaining} to generate adversarial samples to fool the classifier network. This type of input purturbation is called adversarial perturbation, i.e., it is designed to increase the loss via gradient ascent in the input space. In out-of-distribution detection literature, several works utilize the opposite perturbation to the input data using the gradient of the maximum softmax score of the predicted label attained from pre-trained network~\cite{liang2018enhancing,Hsu_2020_CVPR,li2021ecacl,liu2021domain}. In VAD literature, however, there are few efforts in utilizing input perturbations, as we can not directly utilize the softmax outputs of some pre-trained network. %

\section{Method}

\subsection{Object Extraction \& Spatio-Temporal Context Cube Construction}
The basic outline of our proposed framework is presented in Fig.~\ref{fig: pipeline-structure}. Following the practice in~\cite{yu2020cloze}, given the time $t$ current frame, we utilize a pretrained object detector to get all the bounding boxes and extract the corresponding foregrounds. Then we resize all the foreground patches to the identical spatial size to facilitate subsequent modeling. Next, we extract foregrounds for each extracted object foreground with the same location $T$ frames backward and $T$ frames forwards. In this way, we construct a spatial-temporal cube of length $2T+1$ for this object at the current frame $t$, which we name as Spatio-Temporal Context Cube (STCC). %

\subsection{Spatio-Temporal Auto-Trans-Encoder}
\captionsetup{width=0.48\textwidth}
\begin{figure}[t]
   \centering
      \includegraphics[width=0.7\linewidth]{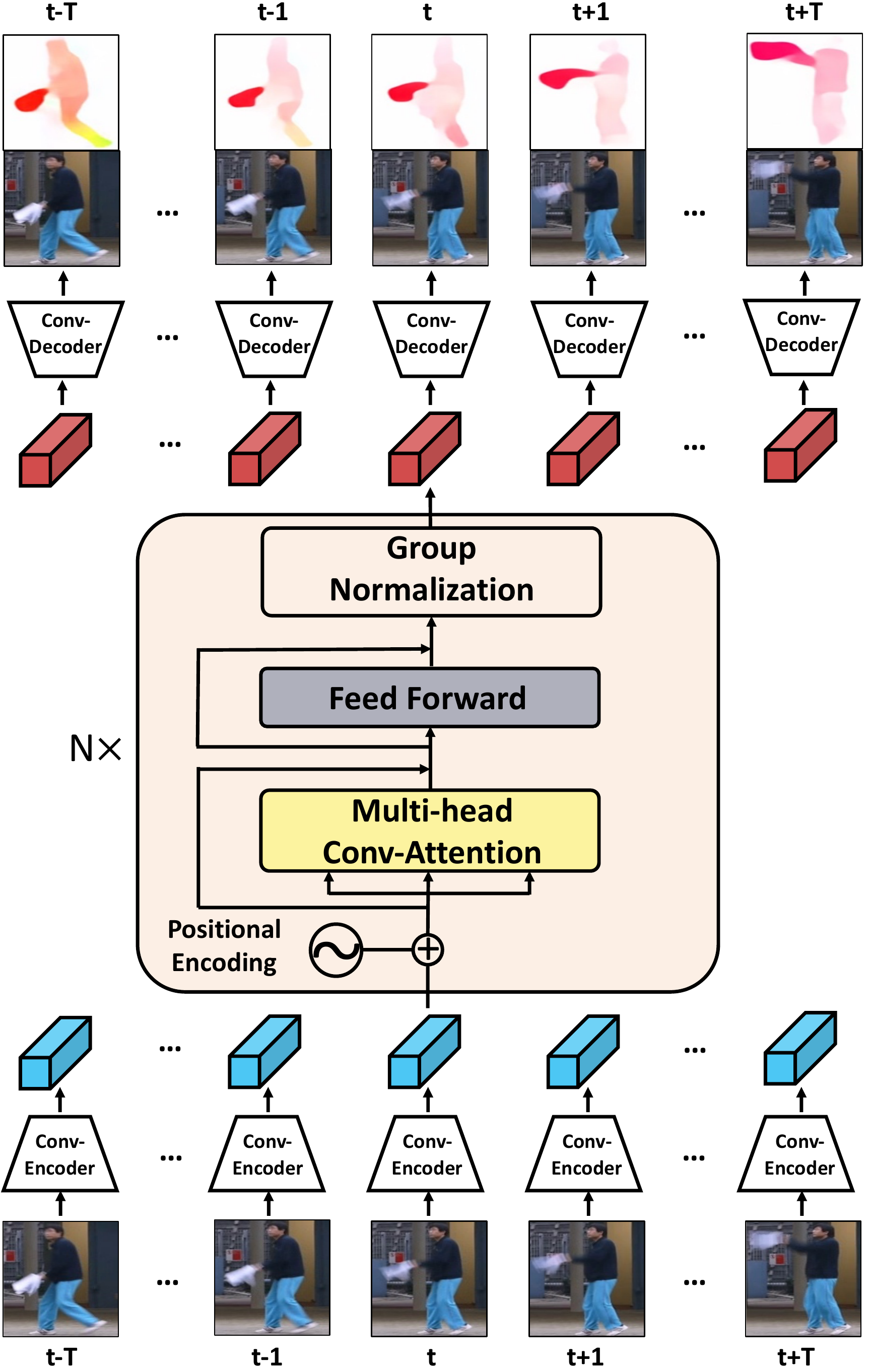}
      \caption{The overall architecture of STATE.} %
   \label{fig: STATE-structure}
\end{figure}

\subsubsection{STATE Architecture Overview}
The whole architecture of STATE is shown in Fig.~\ref{fig: STATE-structure}. Given STCC at time $t$, 
\small
$$\cX_t = \{X_{t-T},\cdots, X_{t+T}~|~X_i\in \RR^{H\times W\times C}, t-T\le i\le t+T \},$$
\normalsize

where $H,W,C$ symbols the height, the width, and the channel number of the input patch respectively, we first let them pass through a convolutional encoder $\mathcal{E}$ to generate representative down-sized feature maps. The same encoder is applied to all the patches of different positions, which gives rise to
\small
$$\{Z_{t-T}, \cdots, Z_{t+T}~|~Z_i\in \RR^{h\times w \times d}, t-T\le i\le t+T\},$$
\normalsize
where $h,w,d$ are the heigth, width and channel number of the downsampled feature maps. Then the feature maps are sent into our uniquely designed transformer-based attention stacks, where feature maps of different positions relate and fuse each other. The outputs of the attention stacks have the same dimension size as their inputs, which we denoted as
\small
$$\{\tilde{Z}_{t-T},\cdots, \tilde{Z}_{t+T}~|~\tilde{Z}_i\in \RR^{h\times w \times d}, t-T\le i\le t+T\}.$$
\normalsize
Eventually these feature maps, after information exchanging, are fed into a convolutional decoder $\mathcal{D}$ which performs spatial upsampling, giving the reconstruction output set
\small
$$\tilde{\cX}_t = \{\tilde{X}_{t-T}, \cdots, \tilde{X}_{t+T}~|~\tilde{X}_i\in \RR^{H\times W \times C'}, t-T\le i\le t+T\}.$$
\normalsize
Here $C'=C=3$ if the model is for raw frame reconstruction , and $C'=2$ if it is for motion reconstruction.

\subsubsection{Attention Stack}

\paragraph{Overview}
 Our attention stack (the boxed part in Fig.~\ref{fig: STATE-structure}) is composed of $N$ layers, and each layer consists of two sub-layers. The initial sub-layer is a multi-head learnable convolutional attention module, and the second is a position-wise 2D convolutional feedforward network with $3\times 3$ conv-filter and Leaky ReLU activation function~\cite{maas2013rectifier}. We adopt residual connection~\cite{he2016deep} throughout each of the sub-layers, and Group Normalization~\cite{wu2018group} is added after the second sub-layer. To facilitate the residual connections, all the sub-layers in the attention stack produce the outputs of channel dimension $d=128$, which is the same as the input feature maps.

\paragraph{Query, Key, Value}
Similar to the transformer~\cite{vaswani2017attention} in NLP, an attention function for a sequence of feature maps can also be narrated as mapping a set and a query of key-value pairs to a corresponding output. Here the query, keys, values, and the output are all three-dimensional tensors. Given the input sequential feature maps 
\small
$$\{Z_{t-T},\cdots, Z_{t+T}~|~Z_i\in \RR^{h\times w \times d}, t-T\le i\le t+T\},$$
\normalsize
we apply convolutional neural networks $W_Q$ and $W_{KV}$ to bring about the query map and the paired key-value map 
\small
$$\{(Q_i, K_i, V_i) \in \RR^{h\times w \times d}~|~t-T\le i\le t+T\}.$$
\normalsize

\paragraph{Positional Encoding}
Different from the original transformer~\cite{vaswani2017attention} which adds positional encodings (PE) to the input embeddings, we choose to add positional encoding to the query map $Q$ and the key map $K$ within each attention stack. %
We propose to use Learnable PE: a learnable tensor embedding of shape $(2T+1,d)$. The advantage of learnable positional encoding is that the positional relationships can be learned in a data-driven way. We use the same positional encoding for different spatial coordinates.

\paragraph{Multi-Head Learnable Convolutional Attention}

\begin{figure}[t]
  \centering
      \includegraphics[width=0.8\linewidth]{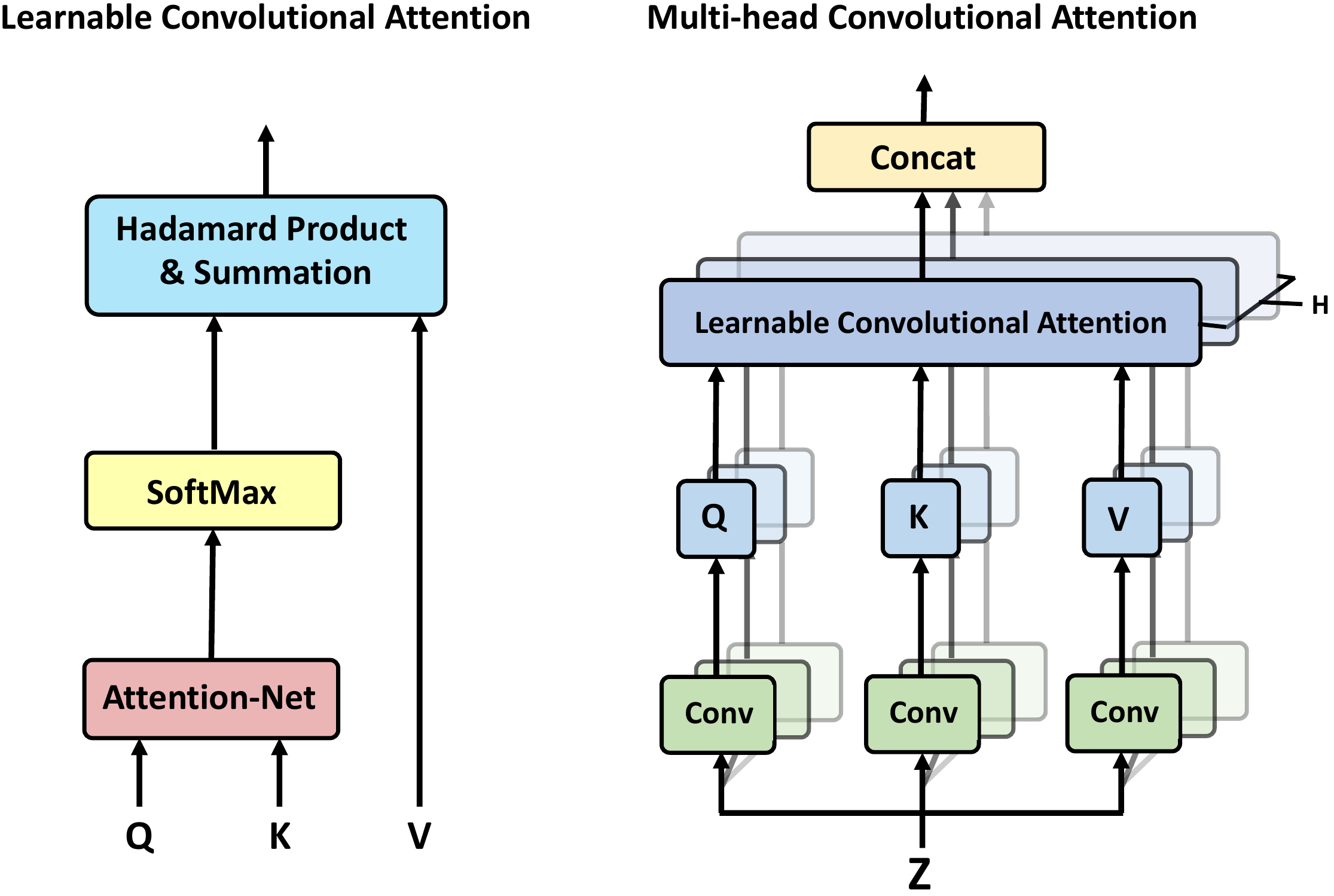}
      \caption{(left) Learnable Convolutional Attention. (right) Multi-Head Convolutional Attention.}
  \label{fig: transformer-detail}
\end{figure}

In each attention stack layer, the key component is the multi-head learnable convolutional attention module (Fig.~\ref{fig: transformer-detail}), which enables feature maps from different frames (positions) to attend to each other and form more discriminative representations with the temporal relationship.
To generate the attention map, we introduce a novel additional network called ``Attention-Net'' $\mathcal{F}$, a one-layer CNN with a $3\times 3$ conv-filter and a Leaky ReLU activation function following. To compute the attention of patch $X_i$ and $X_j$, we first concatenate the query map at $i$-th position $Q_i$ and the key map at $j$-th position $K_j$ along the channel dimension. Subsequently, we apply $\mathcal{F}$ to the concatenation to generate an attention map
\small
\begin{equation*}
   A_i^j = \mathcal{F}\left(\text{concat}\left(Q_i,K_j\right)\right) \in \RR^{h\times w\times 1}.
\end{equation*}
\normalsize
After iterating $j$ from $t-T$ to $t+T$, we get all the attention maps of $X_i$: $\{A_i^{t-T}, A_i^{t-T+1}, \cdots, A_i^{t+T}\}$. Concatenating them we obtain $A_i \in \RR^{(2T+1)\times h \times w\times 1}$. To normalize the values of $A_i$ to weights in $[0,1]$, we apply softmax operation w.r.t. the third dimension:
\small
\begin{equation*}
   A_i[j,...] = \frac{\exp(A_i[j,...])}{\sum_{l=t-T}^{t+T} \exp (A_i[l,...])}, j=1, \cdots, 2T+1.
\end{equation*}
\normalsize
Considering that the pixel-wise attention scores are induced using a learnable convolutional attention generation network, we call our particular attention ``Learnable Convolutional Attention'' as depicted in the left part of Fig.~\ref{fig: transformer-detail}.
Finally the output can be expressed as the summation of 
\small
\begin{equation*}
   \tilde{Z}_i = \sum_{j=t-T}^{t+T}A_i^j\odot  V_j,
\end{equation*}
\normalsize
where $\odot$ denotes Hadamard product (element-wise product). There is broadcasting in multiplication as the last dimension of $A_i^j$ is 1 while the last dimension of $V_j$ is $d$.

In practice, to jointly tackle information from dissimilar representation spaces at different positions, we adopt multi-head attention as indicated in the right part of Fig.~\ref{fig: transformer-detail}:
\small
\begin{equation}\label{eq: concat}
   \text{MultiHead}(Q,K,V) = \text{Concat}(\text{head}_1,\cdots, \text{head}_H),
\end{equation}
\normalsize
where 
\small
\begin{align*}
   \text{head}_h &= \tilde{Z}^{(h)} \\
   &=\text{Attention}\left(Q^{(h)},K^{(h)},V^{(h)}\right) \\
   &= \text{Attention}\left(W_Q^{(h)}* Z, W_{KV}^{(h)}* Z,W_{KV}^{(h)}* Z\right).
\end{align*}
\normalsize
Here $H$ is the number of heads, $1\le h\le H$ and $*$ denotes the 2D convolution operation. For every $h$, the channel dimension of $Q^{(h)},K^{(h)},V^{(h)}$ is $\frac{d}{H}$ and the concatenation in~\eqref{eq: concat} is over the channel dimension of the feature maps.
\captionsetup{width=0.4\textwidth}
\subsubsection{Convolutional Encoder \& Decoder}
The convolutional encoders and decoders in our STATE model take the responsibility of reducing spatial dimensions and extracting deep features. The encoder $\mathcal{E}$ consists of 3 layer blocks. The first block is a 2-layer convolutional net equipped with 3$\times$3 conv filters and padding 1. ReLU and Batch Normalization is employed in between the layers. There are additional max-pooling layers in the next two blocks to reduce spatial dimension. The decoder $\mathcal{D}$ is symmetric to the encoder $\mathcal{E}$ in architecture with the first two blocks performing deconvolution and the last block merely alternating channel dimension. %

\subsection{Training}
During training, each object extracted on each frame generates an STCC. Therefore we form a dataset of STCCs on which we train our STATE model. Suppose the batch size is $N$ and the context length is $T$. For $1\le i \le N$, the $i$-th STCC can be represented as $\cX^{(i)} = [X_{1}^{(i)},\cdots,X_{2T+1}^{(i)}]^\top$. The loss function for raw pixel reconstruction is :
\small
\begin{align*}
   \cL_r &= \frac{1}{N}\sum_{i=1}^N\norm{\text{STATE-raw}\left(\cX^{(i)}\right) - \cX^{(i)}}_p^p \\
   &= \frac{1}{N}\sum_{i=1}^N \sum_{j=1}^{2T+1} \norm{\tilde{X}^{(i)}_{\text{raw}, j}- X^{(i)}_{j}}_p^p,
\end{align*}
\normalsize
where $\tilde{X}_{\text{raw}}$ denotes the output of raw branch of our STATE model during the training, $\norm{\cdot}_p^p$ denotes $p$-norm.
The reconstructions of raw frame patches depict anomalous events with unusual or abnormal appearance (e.g., bicycles, vehicles, etc.). Nevertheless, a considerable proportion of anomalies in videos come from irregular human actions (e.g., fast running, throwing objects, fighting, etc.). To better handle such cases, we add motion reconstruction constraint loss. In practice, for two consecutive frames, we apply a pre-trained FlowNet 2.0~\cite{ilg2017flownet} model to calculate the former frame's optical flow maps as motion information. Similar to $L_r$, the motion branch loss is designed as follows:
\small
\begin{align*}
   \cL_m &= \frac{1}{N}\sum_{i=1}^N\norm{\text{STATE-motion}\left(\cX^{(i)}\right) - \text{FlowNet}\left(\cX^{(i)}\right)}_p^p \\
   &= \frac{1}{N}\sum_{i=1}^N \sum_{j=1}^{2T+1} \norm{\tilde{X}^{(i)}_{\text{flow}, j}- \text{FlowNet}\left(X^{(i)}_{j}\right)}_p^p,
\end{align*}
\normalsize
where $\tilde{X}_{\text{flow}}$ denotes the output of motion branch of our STATE model during the training, and the weights of the FlowNet is fixed. The overall training loss is the  sum of the two losses.

\subsection{Reconstruction-based Input Perturbation}
During the testing phase, suppose there are $M$ objects detected at frame $t$ by the pretrained object detector, we can extract $M$ STCCs $\cY_t^{(1)}, \cdots, \cY_t^{(M)}$ like the training process. We denote the extracted corresponding optical flow map cubes as $\cO_t^{(1)}, \cdots, \cO_t^{(M)}$. For each STCC $\cY_t^{(m)} (1\le m\le M)$ we first apply input perturbation technique to the video sequences using the following formula:
\small
\begin{equation}\label{eq: perturb}
\hat{\cY}^{(m)} = \cY^{(m)} - \eta\text{sign}\left(\nabla_{\cY^{(m)}} \cL_p \left(\cY_t^{(1)}, \cdots, \cY_t^{(M)}\right)\right),
\end{equation}
\normalsize
where sign($\cdot$) denotes the element-wise sign of the tensor, $\eta$ is the perturbation magnitude and 
we utilize the gradient of reconstruction error w.r.t. the testing input raw frames, which can be calculated as: 
\small
\begin{align*}
   \nabla_{\cY^{(m))}}&\cL_p \left(\cY_t^{(1)}, \cdots, \cY_t^{(M)} \right)\\
    =& \frac{1}{M} \bigg(\nabla_{\cY^{(m)}} \norm{\cY^{(m)} - \text{STATE-raw}\left(\cY^{(m)}\right)}_p^p \\
   &+ \nabla_{\cY^{(m)}}\norm{\cO^{(m)} - \text{STATE-flow}\left(\cY^{(m)}\right)}_p^p \bigg).
\end{align*}
\normalsize
The aim of ~\eqref{eq: perturb} is to reduce the reconstruction error given by our STATE model in test time via one-step sign gradient descent on the input frames. We discover that by perturbing the same magnitude, the normal frames tend to reduce their reconstruction error more than anomalous frames. %
\captionsetup{width=0.9\textwidth}
\begin{table*}[t]
   \centering
   \caption{AUROC (\%) comparison with the state-of-the-art methods for VAD on the benchmark datasets.}
   \begin{tabular}{ccccc}
   \toprule[1pt]
   Method Type       &   Method      &   CUHK Avenue     &  ShanghaiTech    \\ 
   \midrule
   \multirow{10}{*}{Reconstruction-based Methods}& ConvAE~\cite{hasan2016learning} (2016)& 80.0& 60.9   \\
                                                & ConvLSTM-AE~\cite{luo2017remembering} (2017)& 77.7& N/A       \\
                                                & Recurrent VAE~\cite{yan2018abnormal} (2019)& 79.6& N/A        \\
                                                & AE+PDE~\cite{abati2019latent} (2019)& N/A& 72.5        \\
                                                & MemAE~\cite{gong2019memorizing} (2019)& 83.3& 71.2       \\
                                                & AMC-AE~\cite{nguyen2019anomaly} (2019)& 86.9& N/A       \\ 
                                                & CDAE~\cite{chang2020clustering} (2020)& 86.0& 73.3 \\
                                                & DGN~\cite{saypadith2021video} (2021) & 86.8& 73.0  \\ \hline
   \multirow{8}{*}{Prediction-based Methods}    & ST-CAE~\cite{zhao2017spatio} (2017)& 80.9& N/A \\
                                                & FramePred~\cite{liu2018future} (2018)& 84.9& 72.8\\ 
                                                & Conv-VRNN~\cite{lu2019future} (2019)& 85.8& N/A \\
                                                & Attention+Prediction~\cite{zhou2019attention} (2019)& 86.0& N/A \\
                                                & MNAD~\cite{park2020learning} (2020)& 88.5& 70.5 \\
                                                & VEC~\cite{yu2020cloze} (2020)& 89.6& 74.8 \\
                                                & VPC~\cite{liu2021deep} (2021)& 85.4& N/A  \\  
                                                & Memory Consistency~\cite{cai2021appearance} (2021)& 86.6& 73.7 \\
                                                & STCEN~\cite{hao2022spatiotemporal} (2022)& 86.6& 73.8 \\ \hline  
   \multirow{4}{*}{Hybrid Methods}              & AnoPCN ~\cite{ye2019anopcn} (2019)& 86.2&73.6 \\
                                                & Skeleton-Trajectories~\cite{morais2019learning} (2019)& N/A& 73.4 \\
                                                & Prediction+Reconstruction~\cite{tang2020integrating} (2020)& 85.1& 73.0\\ 
                                                & sRNN~\cite{8851288} (2021)& 81.7& 68.0 \\ \hline

   \multirow{2}{*}{\bf Ours}                    & STATE & 89.8 & 73.7 \\                                                                     
                  & STATE w/ perturbation & {\bf 90.3}&  {\bf 77.8} \\
   \bottomrule[1pt]
   \end{tabular}
   \label{table: main}
\end{table*}
\subsection{Anomaly Score}
In accordance with previous reconstruction-based methods, we employ the reconstruction error with perturbed input as the anomaly score during testing. Specifically, given the $m$-th STCC at time $t$ $\cY_t^{(m)}$, we adopt the weighted summation of the standardized scores of the raw patch and the corresponding motion map:
\small
\begin{equation*}
   \cS\left(\cY_t^{(m)}\right) = w_r \cdot \frac{\cS_r\left(\cY_t^{(m)}\right) - \overline{S_r}}{\sigma_r} + w_m \cdot \frac{\cS_m\left(\cY_t^{(m)}\right) - \overline{S_m}}{\sigma_m},
\end{equation*}
\normalsize
where
\small
\begin{align*}
   \cS_r\left(\cY_t^{(m)}\right) &= \norm{\text{STATE-raw}\left(\hat{\cY}^{(m)}\right) - \hat{\cY}^{(m)}}_p^p \\
                      &=  \sum_{j=1}^{2T+1} \norm{\tilde{\hat{Y}}^{(m)}_{\text{raw},j} - \hat{Y}^{(m)}_{j}}_p^p, \\
   \cS_m\left(\cY_t^{(m)}\right) &= \norm{\text{STATE-motion}\left(\hat{\cY}^{(m)}\right) - \cO^{(m)}}_p^p \\
   &=  \sum_{j=1}^{2T+1} \norm{\tilde{\hat{Y}}^{(m)}_{\text{flow},j} - \text{FlowNet}(Y^{(m)}_{j})}_p^p, \\
\end{align*}
\normalsize
are the reconstruction errors of raw maps and optical flow maps, and $\overline{S_r}, \sigma_r, \overline{S_m}, \sigma_m$ are the means and standard deviations of the corresponding scores of the raw branch and the motion branch in the training. $w_r$ and $w_m$ are the tunable weights. Here $\tilde{\hat{Y}}_{\text{raw}}$ denotes the output of raw branch and $\tilde{\hat{Y}}_{\text{flow}}$ denotes the output of motion branch in testing using perturbed inputs. Finally the frame score can be computed as the maximum of all the STCCs' scores:
\small
\begin{equation}\label{eq: final-score}
   \cS(\text{Frame $t$}) = \max_{1 \le m \le M} \cS\left(\cY_t^{(m)}\right).
\end{equation}
\normalsize

\section{Experiments}
\subsection{Datasets}
We evaluate our methods on two VAD benchmark datasets CUHK Avenue~\cite{lu2013abnormal} and ShanghaiTech Campus dataset~\cite{8851288}.

\subsection{Experimental Setup} 
In the process of bounding box generation, we use a Cascade R-CNN~\cite{cai2018cascade} network pre-trained on the COCO dataset as the object detecter. We follow the setting in~\cite{yu2020cloze} for the ROI extraction algorithm, setting all the confidence thresholds and overlapping ratios as the same as in ~\cite{yu2020cloze}. For spatio-temporal context cube reconstruction, we set the patch size $H=W=32$. We set context length $T=3$ for STCC construction. As to our proposed model STATE, for both datasets, the hidden size of the three-layer blocks in the convolutional encoder are $32, 64, 128$, and the decoder is symmetric to the encoder with the same hidden size. We equip $W_Q$ and $W_{KV}$ with $3\times 3$ conv-filters and Leaky ReLU activation. We employ $4$ attention heads, which means that the feature map channel number in each head is $32$. We set attention stack number $N=3$ and epoch number to be $20$. In our experiments, we find selecting norm factor $p=2$ is enought to exhibit satisfactory performance. During training, ADAM~\cite{kingma2014adam,wang2021adapting} optimizer with learning rate $0.001$ and $\epsilon$ value $1e-7$ are adopted for the model optimization.  During the testing phase, we choose the perturbation magnitude $\eta = 0.002$ for Avenue and $\eta = 0.005$ for ShanghaiTech. When computing anomaly scores, we set $(0.3, 1)$ for Avenue and $(1, 1)$ for ShanghaiTech. To evaluate anomaly detection performance, we adopt the standard Area Under the Receiver Operating Characteristic curves (AUROC) computed via frame-level criteria. %

\subsection{Result}
We compare our method with 21 SOTA methods in VAD, among which 8 are reconstruction-based, 9 are prediction-based and 4 are hybrid methods. The results are summarized in Tab.~\ref{table: main}. We can observe from Tab.~\ref{table: main} that our method outperforms all the existing methods on both the datasets, reaching AUROC of 90.3\% on Avenue, and 77.8\% on ShanghaiTech. Moreover, compared to the methods based on reconstruction, our method achieves significant performance gain with up to 4.4\% AUROC improvement on Avenue, and 5.5\% AUROC improvement on ShanghaiTech. It is also noticeable that without input perturbation, our STATE model can still achieve SOTA results on Avenue and very competitive performance on ShanghaiTech dataset.

\section{Conclusion}
We have introduced an object-level reconstruction-based framework for video anomaly detection. We put forward a novel transformer-based spatio-temporal auto-encoder for reconstruction. We further incorporate input perturbation technique into reconstruction error and successfully mitigate the overfitting problem of reconstruction-based method. Experiments validate the effectiveness of our approach. %

\section*{Acknowledgment}
Research was sponsored by the DEVCOM Analysis Center and was accomplished under Cooperative Agreement Number W911NF-22-2-0001. The views and conclusions contained in this document are those of the authors and should not be interpreted as representing the official policies, either expressed or implied, of the Army Research Office or the U.S. Government. The U.S. Government is authorized to reproduce and distribute reprints for Government purposes notwithstanding any copyright notation herein.

\bibliographystyle{IEEEtran}
\bibliography{vad}

\end{document}